%% file: tal.tex
\documentclass[10pt,twocolumn,letterpaper]{article}

\usepackage{cvpr}              %

\usepackage{graphicx}
\usepackage{amsmath}
\usepackage{amssymb}
\usepackage{booktabs}

\newcommand{\dt}[1]{\fontsize{7pt}{0.1em}\selectfont (#1)}
\newcommand{\bd}[1]{\textbf{#1}}

\newlength\savewidth\newcommand\shline{\noalign{\global\savewidth\arrayrulewidth
  \global\arrayrulewidth 1pt}\hline\noalign{\global\arrayrulewidth\savewidth}}
\newcommand{\tablestyle}[2]{\setlength{\tabcolsep}{#1}\renewcommand{\arraystretch}{#2}\centering\small}

\usepackage{enumitem}  %
\usepackage{multirow}
\usepackage{tabularx}
\usepackage{booktabs}
\usepackage{xcolor}
\usepackage{pifont}
\usepackage{bm} 
\newcommand{\cmark}{\ding{51}}%

\usepackage[pagebackref,breaklinks,colorlinks]{hyperref}

\usepackage[capitalize]{cleveref}
\crefname{section}{Sec.}{Secs.}
\Crefname{section}{Section}{Sections}
\Crefname{table}{Table}{Tables}
\crefname{table}{Tab.}{Tabs.}

\def\cvprPaperID{549} %
\def\confName{CVPR}
\def\confYear{2022}

\begin{document}

\title{RCL: Recurrent Continuous Localization for Temporal Action Detection}

\author{Qiang Wang, Yanhao Zhang, Yun Zheng, Pan Pan\\
DAMO Academy, Alibaba Group\\
{\tt\small \{qishi.wq, yanhao.zyh, zhengyun.zy, panpan.pp\}@alibaba-inc.com}
}
\maketitle

\input{abstract}

\input{introduction}

\input{related}

\input{methodology}

\input{experiments}

\input{conclusion}

\clearpage
{\small
\bibliographystyle{ieee_fullname}
\bibliography{tal}
}

\end{document}

%% file: abstract.tex
\begin{abstract}

Temporal representation is the cornerstone of modern action detection techniques.
State-of-the-art methods mostly rely on a dense anchoring scheme, where anchors are sampled uniformly over the temporal domain with a discretized grid, and then regress the accurate boundaries.
In this paper, we revisit this foundational stage and introduce Recurrent Continuous Localization (RCL), which learns a fully continuous anchoring representation.
Specifically, the proposed representation builds upon an explicit model conditioned with video embeddings and temporal coordinates, which ensure the capability of detecting segments with arbitrary length.
To optimize the continuous representation, we develop an effective scale-invariant sampling strategy and recurrently refine the prediction in subsequent iterations.
Our continuous anchoring scheme is fully differentiable, allowing to be seamlessly integrated into existing detectors, \textit{e.g.}, BMN~\cite{bmn} and G-TAD~\cite{gtad}. 
Extensive experiments on two benchmarks demonstrate that our continuous representation steadily surpasses other discretized counterparts by $\mathtt{\sim}2\%$ mAP.
As a result, RCL achieves $52.92\%$ mAP@0.5 on THUMOS14 and $37.65\%$ mAP on ActivtiyNet v1.3, outperforming all existing single-model detectors.

\end{abstract}

%% file: introduction.tex
\section{Introduction}
\label{sec:intro}

Temporal Action Localization (TAL) that localizes temporal boundaries of actions with specific categories in untrimmed videos~\cite{asm,thumos14,anet}, is at the core of several down-stream tasks such as video classification~\cite{slowfast}, video captioning~\cite{zhang2020object} and video editing~\cite{autovideoedit}. 
This challenging problem has been deeply studied in recent years~\cite{rc3d,cdc,bsn,bmn,gtad}, as the large scale variation problems is very serious, causing sophisticated feature designs to capture both local and global information, and thus inspired many extensions such as UNet-like architecture~\cite{pbrnet}, local context~\cite{tcanet}, and proposal-relations~\cite{pgcn,gtad,vsgn}. 
Prior works~\cite{rc3d,talnet} take inspiration from image detection~\cite{fasterrcnn,yolo} and are carried out by densely spanning temporal anchors and predicting their corresponding scores.
Other challenges include the fact that the definition of an action’s temporal boundaries are often ambiguous~\cite{detad}. 
The ambiguity and uncertainty also hinder the convergence for localization optimization, and brings an illogical empirical observation that the classification-based detectors~\cite{bmn,gtad} usually achieve better performance than regression-based methods~\cite{talnet}.

\begin{figure}[t]
\centering
\centering
\subfloat[\centering the overall anchor distribution]{\includegraphics[width=0.49\linewidth]{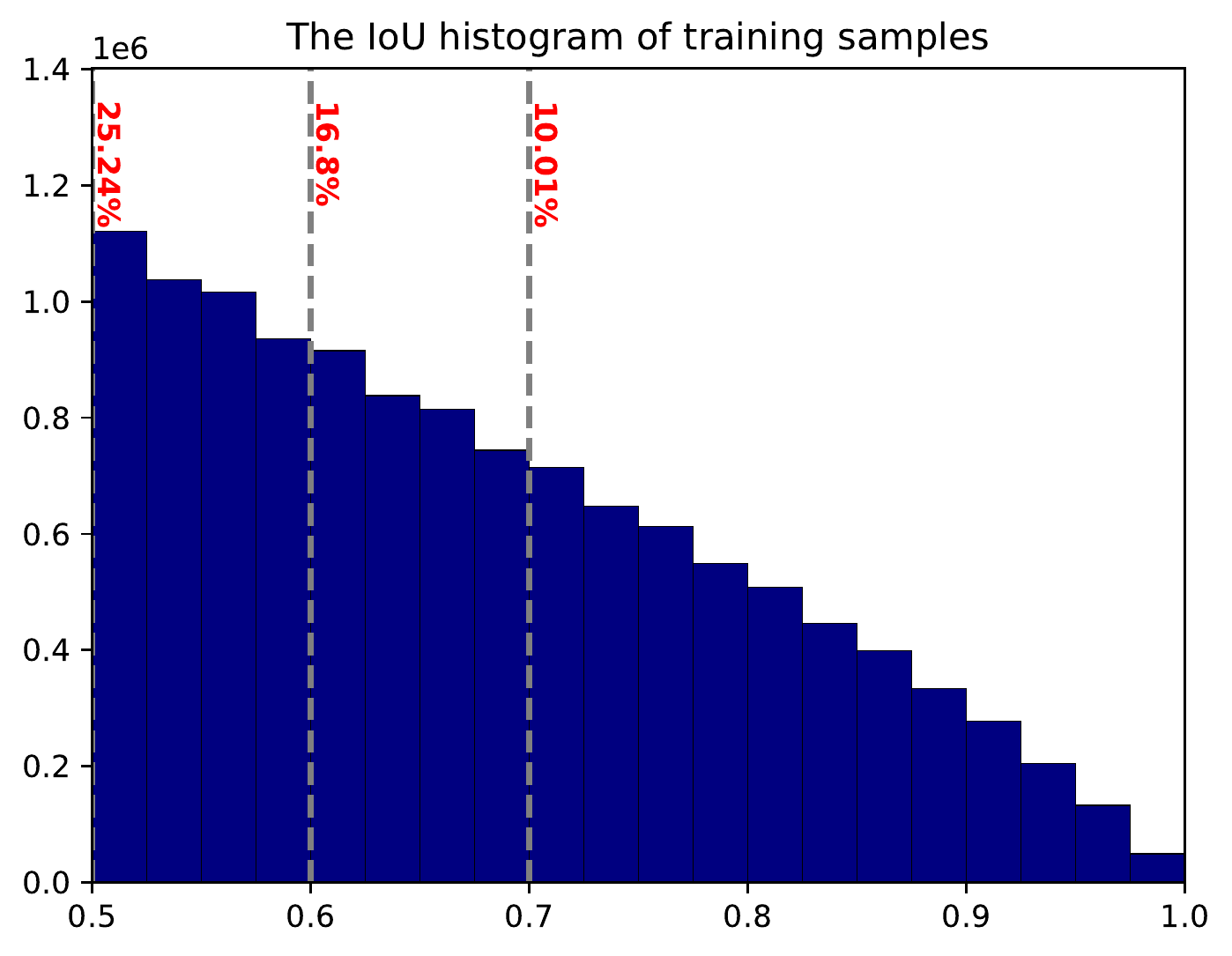}}%
\subfloat[\centering the unbalanced scale distribution]{\includegraphics[width=0.49\linewidth]{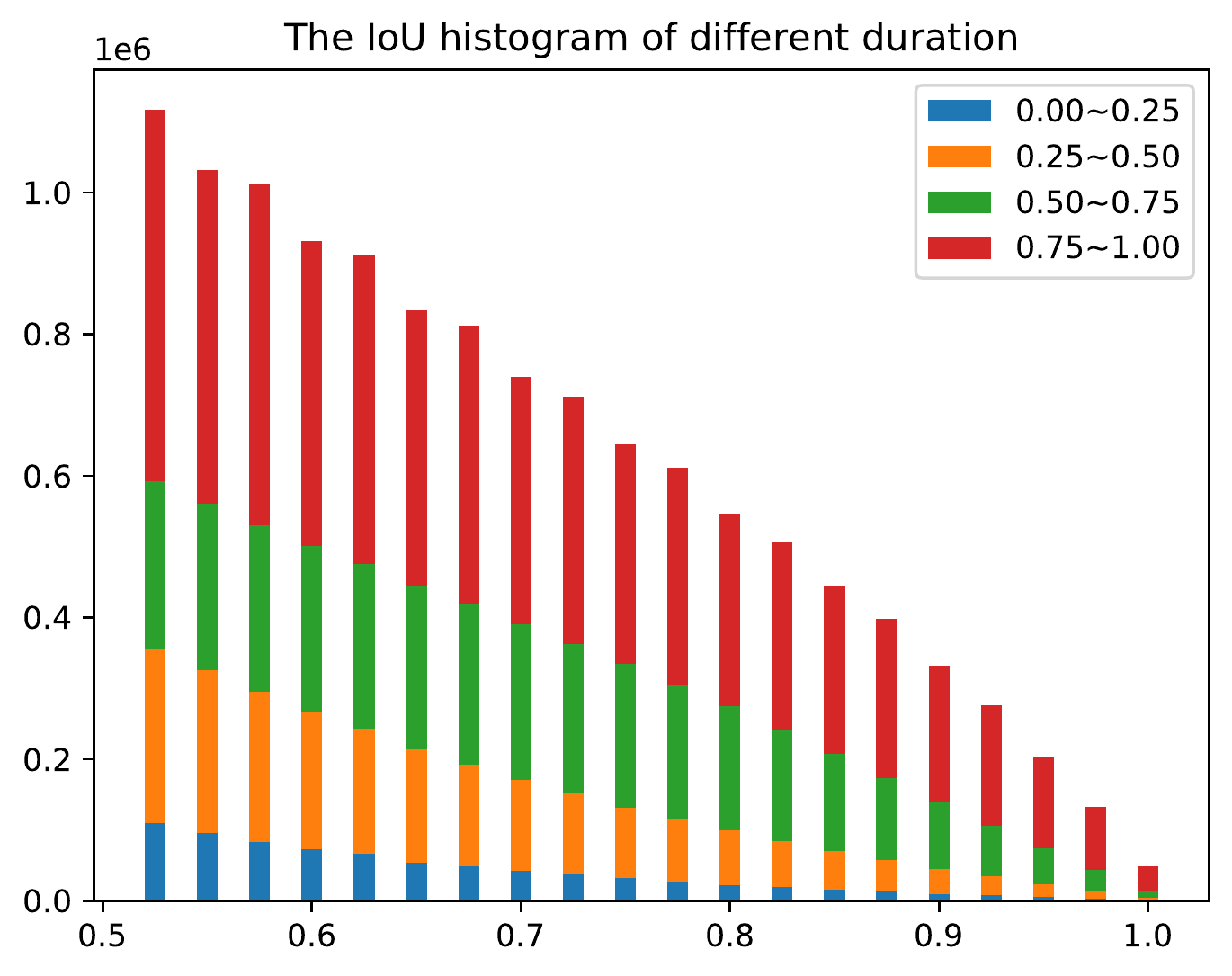}}
\caption{The tIoU histogram of training anchors from BMN~\cite{bmn}. The red numbers are the positive percentage higher than the corresponding tIoU threshold. These anchors mainly cover the long segments, which results in a missing detection for short instances.}
\label{fig:anchor}
\end{figure}

\begin{figure*}[t]
\centering
\centering
\includegraphics[width=0.99\linewidth]{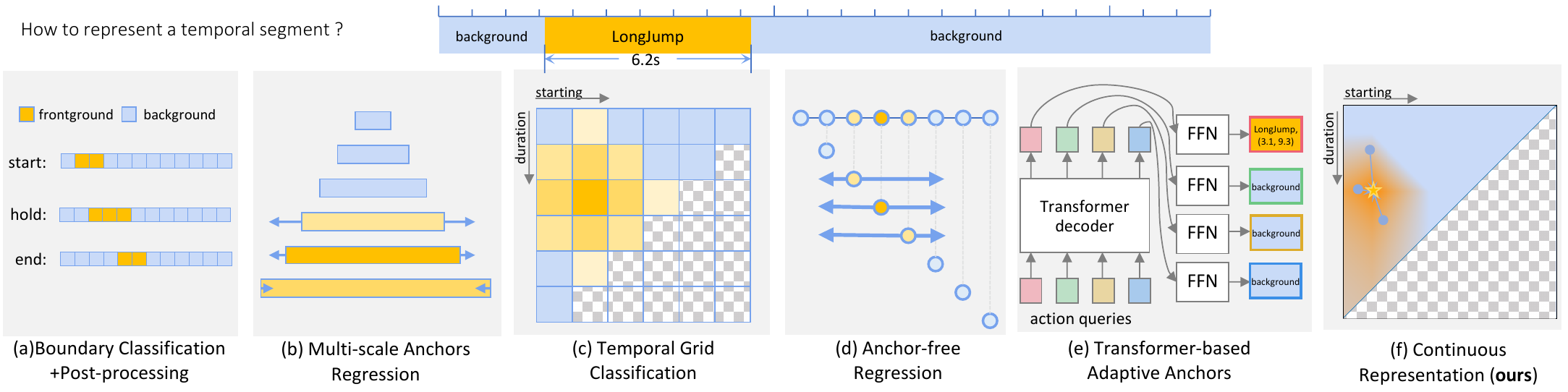}
\caption{The typical temporal representation methods. (a) the bottom-up representation~\cite{bsn,icic}. (b) the multi-scale anchor representation~\cite{rc3d,talnet}. (c) the grid-based representation~\cite{bmn,2dtan}. (d) the anchor-free representation~\cite{afsd}. (e) the transformer-based representation~\cite{rtd} (f) the proposed continuous representation. Best viewed in color.}
\label{fig:cr}
\end{figure*}

While numerous efforts have been made towards solving the above challenges, recent approaches still suffer from a major limitation: they mainly leverage a \textit{discretized} anchoring representation. 
For example, existing bottom-up methods~\cite{cdc,bsn,icic} utilize the discretized boundary classification and a well-tuned post-processing to compose temporal segments, which can not be trained in an end-to-end manner.
Recently, many works utilize the pre-defined temporal anchor to represent the temporal hypothesis, \textit{e.g.},  the sliding-windows paradigm~\cite{shou2016temporal} and the multi-scale anchors~\cite{rc3d,talnet}. 
These methods show excellent performance with faster speed and have the ability to handle large duration segments.
In contrast to representing complete segments, some anchor-free methods, \textit{e.g.}, AFSD~\cite{afsd}, leverage center-point representations to directly regress the start and end time, and the latest studies~\cite{rtd,oadtr} utilize transformer decoder to bi-match the segments with action queries.
In general, different representation methods usually steer the detectors to perform well in different aspects. 
For example, the bottom-up representation is usually more accurate for fine-grained localization.
The anchor-based representation achieves better completeness and is easy to optimize with the tIoU supervision. 
The anchor-free representations avoid the need for an anchoring design and are usually quite efficient.
The transformer has shown powerful abilities with set matching loss from action queries. 
Noticing that different representations and their anchoring optimization are usually heterogeneous, 
but their performances essentially depend on the anchor \textit{distribution} and the \textit{ranking} quality between the anchors. 
As shown in Figure~\ref{fig:anchor}, the discretized anchoring representation~\cite{bmn} can only provide coarse proposals, causing seriously missed detection for short-term segments.

To address this issue, we introduce a novel anchoring representation that is efficient, expressive, and fully continuous, as depicted in Figure~\ref{fig:cr}(f). 
Our key idea is to directly regress confidence scores from continuous anchor points using deep neural networks. 
Thus we can extract precise segments by searching local maximum in the continuous function.

In this work, we present Recurrent Continuous Localization (RCL), an explicit model conditioned with video embeddings and temporal coordinates.
Our approach uses the concept of a Continuous Anchoring Representation (CAR) to achieve high fidelity action detection. 
Unlike common anchor-based detection techniques, which discretize the segments into a regular grid for measurement~\cite{bmn}, we produce an estimation in the continuous field.
The proposed continuous representation can be intuitively understood as a learned position-conditioned classifier for which the confidence scores are jointly determined by the video features and the temporal coordinates.

The proposed RCL can serve as a generic plug-in module into various prevalent temporal action localization frameworks, including BMN~\cite{bmn} and G-TAD~\cite{gtad}. 
Extensive experiments on the THUMOS14~\cite{thumos14} and ActivityNet v1.3~\cite{anet} show that RCL substantially improves various detectors by $2\sim4\%$ mAP. 
In particular, we improve a strong BMN detector by about $1.8\%$ average mAP and $5.9\%$ recall, reaching $37.65\%$ mAP on ActivityNet v1.3. 

 The innovations of this article are as follows:
 \setlist{nolistsep}
 \begin{itemize}
 \item We propose a continuous anchoring representation method, which unifies and extends existing anchor-based detector into a continuous regression problem in 2D coordinates.
 \item To optimize the continuous representation, we develop an effective scale-invariant sampling strategy, which provide accurate ranking scores for short-term segments. 
 \item With an iterative optimization method, our model adaptively focus on target region and provide a refined estimation.
 \item Our model obtain state-of-the-art results in quantitative comparisons on the THUMOS14~\cite{thumos14} and ActivityNet v1.3~\cite{anet} datasets, with $52.92\%$ mAP@0.5, $37.65\%$ mAP, respectively.
 \end{itemize}

%% file: related.tex
\section{Related Work}
\label{sec:related}

Temporal Action localization (TAL) aims to find all segments in an untrimmed video with their location described by 2D temporal coordinates. 
To discriminate action segments from background, intermediate geometric candidates and their corresponding features are required. 
Here we mainly concentrate on the geometric representations, where typical representations used in TAL are illustrated in Figure~\ref{fig:cr} and summarized below.

\noindent\textbf{Bottom-up representation}. 
Early TAL frameworks~\cite{cdc,bsn,icic} involve evaluating the snippet-level probabilities of three action-indicating phases, \textit{i.e.} starting, continuing, and ending; and obtain temporal boundaries via a intensive post-processing step.
They provide an intuitive way to determine a segment by two key moments $(\mathbf{x}_s, \mathbf{x}_e)$. While the heuristic merge operation is usually not fully differentiable, which leads to a inferior performance.

\noindent\textbf{Multi-scale anchor representation}. 
Inspired by anchor-based image detection~\cite{fasterrcnn,yolo}, the first family of anchor-based TAL methods~\cite{rc3d,talnet} typically employ the multi-scale anchor representation and attach an auxiliary boundary regression branch to refine these pre-defined anchors. 
Geometrically, given pre-defined anchors $(\mathbf{a}_{s,k}, \mathbf{a}_{e,k})$, the network simultaneously predicts the confidence score $\mathbf{s}_k$ and the relative offset $(\bigtriangleup \mathbf{a}_{s,k}, \bigtriangleup \mathbf{a}_{e,k})$. 
While the large scale variations in the duration make it challenging to recognize localization boundary~\cite{talnet}. 
The fixed small set of anchors are also less flexible to cover a complexity distribution. 
Cascaded localization strategies~\cite{pbrnet,tcanet} are usually employed to alleviate this issue.

\noindent\textbf{Grid-based representation}. 
In order to further increase the sampling density, a straightforward solution is to densely enumerate all segments and predict the corresponding confidence scores~\cite{sst}.  
\cite{bmn,gtad,2dtan} ingeniously express this enumeration structure in discretized 2D grids, and optimize a 2D heatmap through 2D/3D convolution.
However, due to the squarely growing compute and memory requirements, current methods are only able to handle low resolutions ($256\times 256$ or below). 
With this coarse discretization, the state-of-the-art grid-based TAL approach, BMN~\cite{bmn}, can only obtain $40.2\%$ recall rate for short segments, leading a low-fidelity prediction on ActivityNet v1.3~\cite{anet}.
For an input video with a snippet-level feature size $T_s$, the sample space of grid-based representation is at a scale of $\mathcal{O}(T_s^2)$. 

\noindent\textbf{Anchor-free representation}. 
To reduce the complexity, some recent frameworks~\cite{afsd} use the center point as a simplified representation and directly regress the boundary location. 
Geometrically, a center point is described by a 1-D vector $(x_c)$ and the hypothesis sample space is in the scale of $\mathcal{O}(T_s)$, which is much more tractable. 
The strategy of reducing the sample significantly increases the training difficulty for the regression branch.
Therefore, the anchor-free methods usually achieve inferior accuracy compared to anchor-based method.

\noindent\textbf{Transformer representation}.
More recently, \cite{rtd, oadtr} introduce the transformer architecture~\cite{detr} to directly predict all segments in parallel, which take advantage of the query-key mechanism and utilize a small set of learned action queries as implicit adaptive anchor.

\noindent\textbf{Continuous representation}.
Recent 3D rendering works~\cite{deepsdf,nerf} propose to utilize the continuous signed distance functions to represent 3D shapes and eliminate discretization errors.
LIIF\cite{liif} extends the continuous representation to image coordinate, which can generate arbitrary super-resolution .
Inspired by the above methods, we introduce the continuous representation to the temporal domain.

%% file: methodology.tex
\section{Methodology}
\label{sec:method}

\begin{figure}[t]
\centering
\centering
\includegraphics[width=0.99\linewidth]{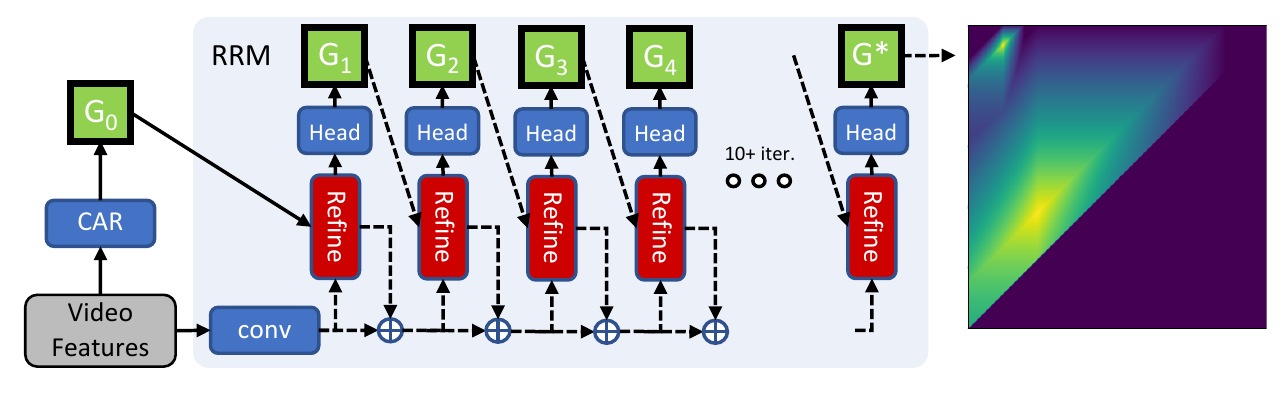}
\caption{RCL consists of 3 main components: (1) A feature encoder that extracts temporal features from an input video. (2) A continuous anchoring representation (CAR) which predicts a continuous confidence map with a scale-invariant sampling strategy. (3) A recurrent refine module (RRM) which updates the confidence map by iteratively refining the uncertain regions.}
\label{fig:framework}
\end{figure}

\noindent\textbf{RCL Overview.} In this section we present RCL, a recurrent continuous localization learning approach. 
Figure~\ref{fig:framework} illustrates the overall pipeline of our method. 
Our method formulates temporal segment as a local maximum in a continuous 2D function $G_{\bm{\theta}}(\mathbf{F}; {\mathbf{x}})$.
We optimize a deep neural network to represent this function, which simultaneously estimates the confidence scores and relative offsets from the snippet features $\mathbf{F}$ and the 2D temporal coordinate $(\mathbf{x}_s,\mathbf{x}_e)$.

The framework takes an untrimmed video frames $\mathbf{V}\in \mathbb{R}^{3\times T\times H\times W}$ as input and estimates all potential temporal segments $\varphi=\{(\mathbf{x}_{s,n},\mathbf{x}_{e,n}, c_n)\}^{N}_{n=1}$ that may contain known actions, and these segments can be represented as key points in continuous 2D confidence maps.
Our method can be distilled down to three stages: (1) video feature extraction, (2) computing continuous 2D confidence maps, and (3) iterative updates, where all stages are differentiable and composed into an end-to-end trainable architecture.

\subsection{Video Feature Extraction}
Following the common practice for temporal action detection approaches~\cite{bsn, bmn,pgcn,gtad}, the video features are offline extracted from the untrimmed video frames using a 3D convolutional network~\cite{i3d,tsn,slowfast,tsp}. 
We adopt the sliding window approach to split the long video into several short snippets, where $\sigma$ is the time interval and $L$ is the length of a snippet.
Our encoder, $\mathbf{F} = f_{\bm{\theta}}(\mathbf{V})$, utilizes spatial average pooling to eliminate the spatial dimension and outputs a compact video feature $f_{\bm{\theta}} : \mathbb{R}^{3\times T\times H\times W} \rightarrow  \mathbb{R}^{D\times T_s}$, where $D$ is the feature dimension and temporal resolution $T_s=\lfloor (T-L+1)/\mathbf{\sigma} \rfloor$ . 
We use the off-the-shelf video recognition models~\cite{tsn, tsp} and freeze the parameters $\bm{\theta}$ of the video feature extractor $f_{\bm{\theta}}$ for training efficiency.

\subsection{Continuous Anchoring Representation}

\begin{figure}[t]
\centering
\centering
\includegraphics[width=0.99\linewidth]{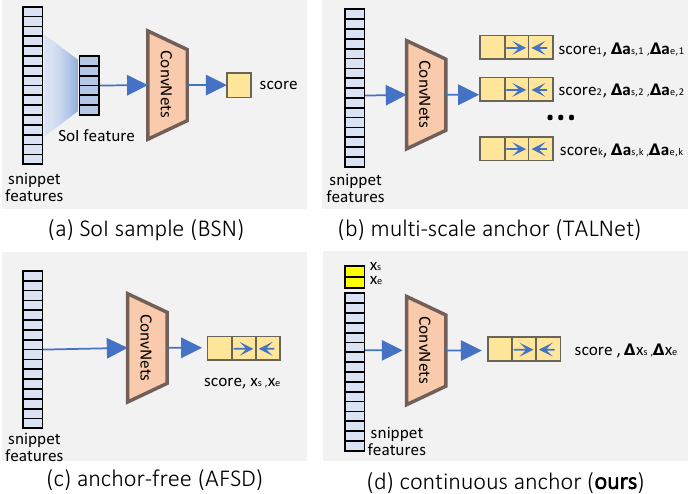}
\caption{In the continuous representation instantiation, the temporal information and the anchor information are concatenated as input. CAR produces the confidence score and the relative offset for any 2D segment query.}
\label{fig:cranchor}
\end{figure}

In this section we present CAR, a continuous representation module, which brings a unified perspective for current geometric representation~\cite{bsn, talnet, afsd}.

As shown in Figure~\ref{fig:cranchor}, the geometric representation for typical temporal anchoring methods can be formulated as three architectures: 

(1) \textbf{The bottom-up methods}~\cite{bsn,icic} first obtain the boundary candidates, and then use the 1D RoI pooling (termed ``SoI'') to estimate all possible combinations. 
Formally, the whole process can be formalized as: 
\begin{equation}
\left\{\begin{array}{c}
G_{\bm{\theta}}(\mathbf{F}; \mathbf{x}) = (p_{\bm{\theta}}(\mathbf{F}; \mathbf{x}), \mathbf{x}_s, \mathbf{x}_e) \\
p_{\bm{\theta}}(\mathbf{F}; \mathbf{x}) = s_{\bm{\theta}}(\mathbf{F}; \mathbf{x}_s) \cdot e_{\bm{\theta}}(\mathbf{F}; \mathbf{x}_e) \cdot q_{\bm{\theta}}(SoI(\mathbf{F}; \mathbf{x}_s, \mathbf{x}_e)),
\end{array}\right.
\end{equation} 
where $s_{\bm{\theta}}$, $e_{\bm{\theta}}$ are two binary classifiers to localize the start time and end time, which are usually implemented with 1D temporal convolution layers. $q_{\bm{\theta}}$ provides confidence score for a proposal and $p_{\bm{\theta}}$ is the fused confidence score.
BSN~\cite{bsn} adopts the cascaded paradigm to determine the start and end location first, and then composes segments via a boundary-sensitive evaluation. 
BMN~\cite{bmn} directly enumerates all candidates, and accelerates SoI through a matrix multiplication, which forms an end-to-end training solution.
However, since the temporal classifier works on the discretized feature $\mathbf{F}$, 
the \textit{smallest representable length} is inversely proportional to the feature size $Duration/T_s$ 
and the size of its sample space is $T_s \cdot(T_s+1)/2$.
To improve the localization accuracy, it is intuitive to rescale the video feature size $T_s$. 
While the computational cost will increase significantly, as analyzed in Section~\ref{sec:ablation}.

(2)\textbf{The multi-scale anchor methods}~\cite{rc3d,talnet} extend image detection, \textit{e.g.} Faster R-CNN~\cite{fasterrcnn}, to temporal action localization. They generate the class-agnostic proposals by jointly classifying and regressing a fixed set of multi-scale anchors $\mathcal{A}= \{(\mathbf{a}_{s,k},\mathbf{a}_{e,k})\}_{k=1}^{K}$ at each location. The coordinate transformations are computed as follows:
\begin{equation}
\left\{\begin{array}{c}
G_{\bm{\theta}}(\mathbf{F}; \mathbf{a}_k) = (p_{\bm{\theta}}(\mathbf{F}; \mathbf{a}), \mathbf{a}_{s,k}^{*}, \mathbf{a}_{e,k}^{*}) \\
\mathbf{a}_{s,k}^{*} = \bigtriangleup_{\bm{\theta}} \mathbf{a}_{s,k} \cdot l_k + \mathbf{a}_{s,k} \\
\mathbf{a}_{e,k}^{*} = \bigtriangleup_{\bm{\theta}} \mathbf{a}_{e,k}  \cdot l_k + \mathbf{a}_{e,k}
\end{array}\right.
\end{equation}
where $\mathbf{a}_{k}$ and $l_k$ are the coordinate and length for the $k$-th anchor. 
Theoretically, the ground-truth segment can be losslessly recovered through the offset regression learning. 
While the design of the anchor itself is a discretized representation, which will cause an imbalance sample problem~\cite{tsi,bsnpp} and make it less flexible.

(3)\textbf{The anchor-free methods}~\cite{afsd} directly predict the confidence score, the center offset and length of time through the center point feature:
\begin{equation}
\left\{\begin{array}{c}
G_{\bm{\theta}}(\mathbf{F}; c_{i}) = (p_{\bm{\theta}}(\mathbf{F}; c_i), c_{i}^{*}-l_{{\bm{\theta}}, i}/2, c_{i}^{*}+l_{{\bm{\theta}},i}/2) \\
c_{i}^{*} = \bigtriangleup_{\bm{\theta}} c_{i} + c_{i} .
\end{array}\right.
\end{equation}
This design makes the system much efficient, but 
the offset optimization will be more difficult, usually resulting in performance degradation.

(4)\textbf{The continuous representation} proposes modeling action segments by maximizing the confidence scores in a 2D function. 
The key difference to the grid-based methods~\cite{bmn,gtad} is that the confidences are defined on a \textit{continuous} temporal domain.
For a given segment $(\mathbf{x}_s, \mathbf{x}_e)$, the continuous function can output the segment's confidence score and relative offset to the closest annotation:
\begin{equation}
\left\{\begin{array}{c}
G_{\bm{\theta}}(\mathbf{F}; \mathbf{x}) = (p_{\bm{\theta}}([\mathbf{F}, \mathbf{x}]; \mathbf{x}), \mathbf{x}_{s}^{*}, \mathbf{x}_{e}^{*}) \\
\mathbf{x}_{s}^{*} = \bigtriangleup_{\bm{\theta}} \mathbf{x}_{s} \cdot (\mathbf{x}_{e}-\mathbf{x}_{s}) + \mathbf{x}_{s} \\
\mathbf{x}_{e}^{*} = \bigtriangleup_{\bm{\theta}} \mathbf{x}_{e}  \cdot (\mathbf{x}_{e}-\mathbf{x}_{s}) + \mathbf{x}_{e}
\end{array}\right.
\end{equation}
where $[,]$ denotes a concatenation operator. Our model is an explicit setting method, which fed the anchor coordinate itself as a \textit{condition} input to constitute the prediction. 
Note that this design differs essentially from the current anchoring schemes (as Figure~\ref{fig:cranchor}) in that every location is associated with a dynamic anchor instead of a set of pre-defined anchors. 
Since it allows arbitrary length, our scheme can better represents the extremely fine-grained segments.
Our experiments show that due to the high-fidelity sample space, we achieve much higher recall than the baseline scheme, please refer to Section~\ref{sec:detad}. 

The continuous design enables more flexible and efficient data sampling space, which shows some appealing properties in Section~\ref{sec:sample}.

\subsection{Sampling Strategy and Feature Alignment}
\label{sec:sample}
\begin{figure}[t]
\centering
\centering
\includegraphics[width=0.99\linewidth]{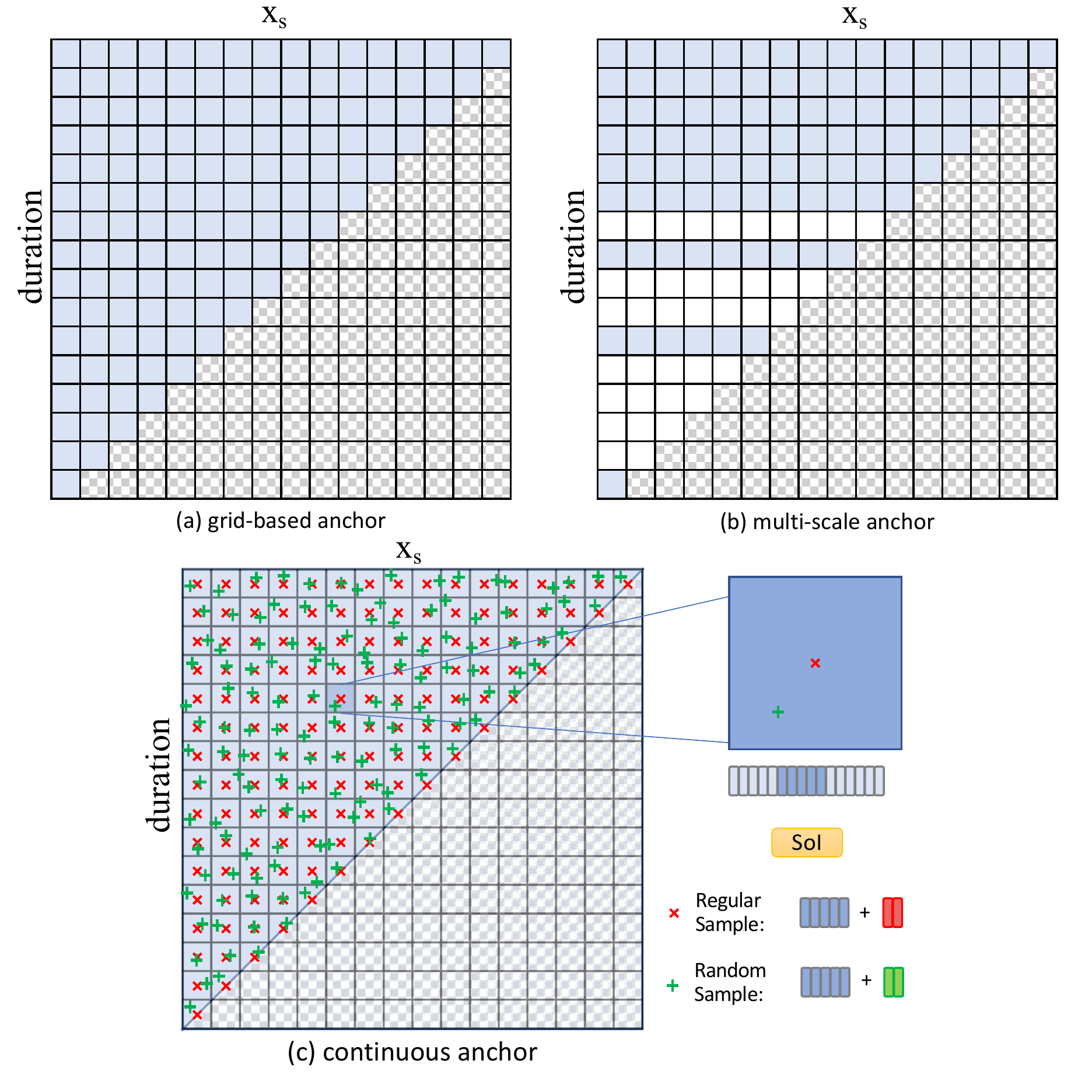}
\caption{An illustration of the sample strategies for (a) grid-based anchor representation, (b) multi-scale anchor representation and (c) the proposed continuous anchor representation. The lightblue box denote the anchors for the discretized representation (a-b). }
\label{fig:sample}
\end{figure}

The proposed representation can be viewed as a continuous extension to the discretized grid representation.
In the actual training process, there are two problems: (1) The continuous representation function contains infinite samples, exhaustive sampling is computationally prohibitive. 
A common solution is to randomly collect some points in each training batch to optimize the overall function~\cite{nerf,deepsdf}.
(2) For each ground-truth segment $(\mathbf{g}_s, \mathbf{g}_e)$, it can be mapped to a point on our 2D axis (Figure~\ref{fig:sample}).
While prior studies~\cite{bmn,tsi} shown that the training samples for different scales are not balanced, the loss terms will be overwhelmed by the long segments. 
For a continuous representation, we can sample on the entire real number domain, which ensures that we can easily control the ratio for different length instances.

To solve the above issues, we propose a scale-invariant sampling strategy. 
(1) \textbf{Regular Grid Samples}: Note that when only sample points in the regular grid centers ($\color{red}\bm{\times}$ in Figure.~\ref{fig:sample} (c)), our continuous representation can degenerate to grid representation~\cite{bmn} (Figure~\ref{fig:sample} (a)).
The valid samples number from 2D-grid is $T_s\cdot (T_s+1)/2$.
(2) \textbf{Random Samples}: To train the continuous function, we random sample $T_s\cdot (T_s+1) / 2$ segments around the regular grid samples ($\color{green}\bm{+}$ in Figure~\ref{fig:sample} (c)).
(3) \textbf{Scale-invariant Samples}: For each ground-truth annotation $(\mathbf{g}_{s,i}, \mathbf{g}_{e,i})$, we sample $n$ points around it, taking its length $l$ as the variance. 
As shown in Figure~\ref{fig:anchor}, for a long segment, there may be hundreds of samples. 
In this case, there are relatively more pairs for long segments.
The number of short-term samples is rare, and the learning of ranks between samples is relatively difficult.
Our balanced sample strategy is very helpful for the rank learning between instances of different lengths. 

In addition, we note that although our method, as a black-box function, can predict a confidence score for arbitrary segment. However, according to Shannon's sampling theorem~\cite{shannon2001mathematical}, our finest input observation is the video frame, and the temporal resolution of our output is still limited. 
Therefore, we keep the minimum output duration at the video frame level, termed SPF (Seconds per Frame). 

\subsection{Recurrent Refine Module}
\label{sec:rrm}
\begin{figure}[t]
\centering
\centering
\subfloat[\centering the offset flow]{\includegraphics[width=0.49\linewidth]{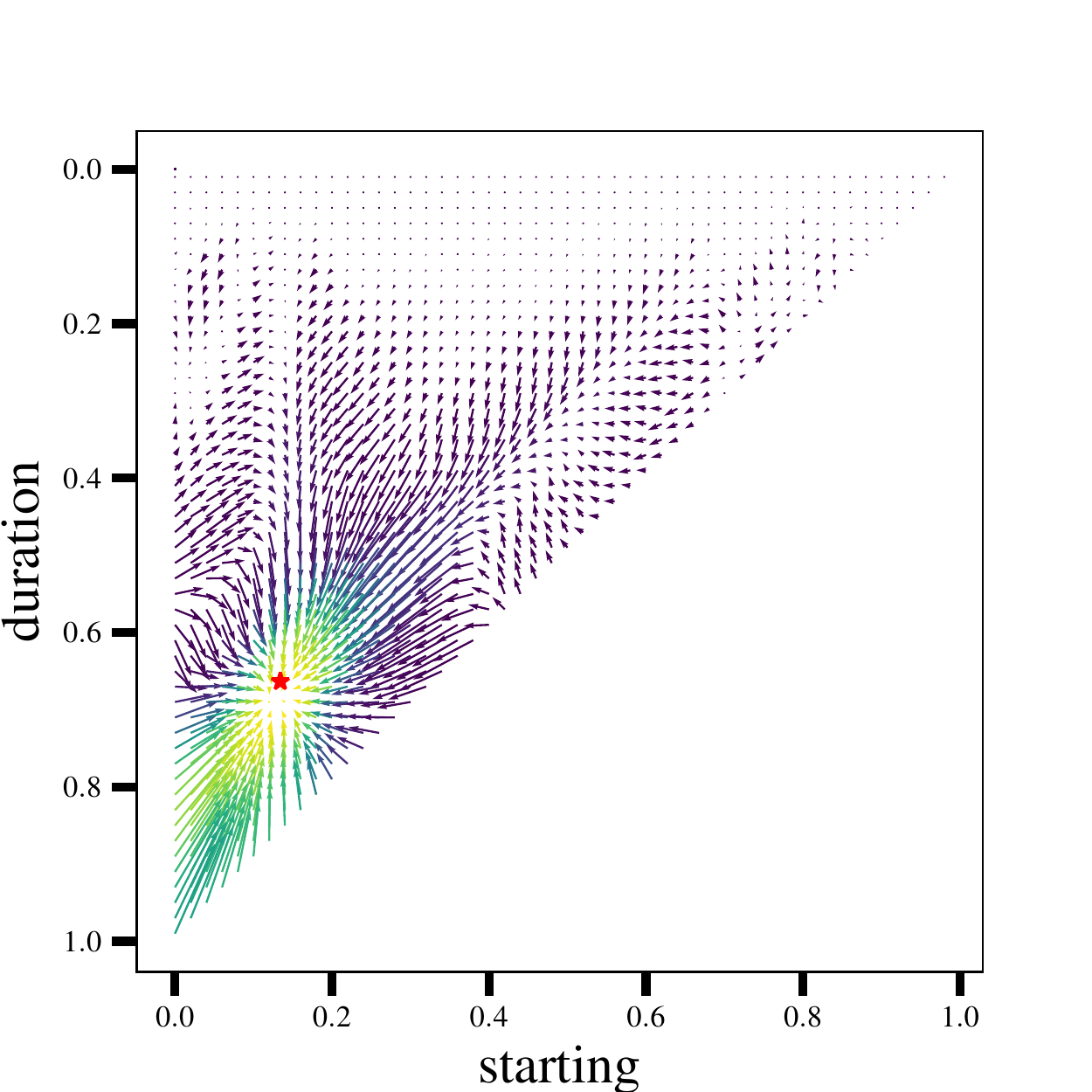}}%
\subfloat[\centering iterative sample distribution]{\includegraphics[width=0.49\linewidth]{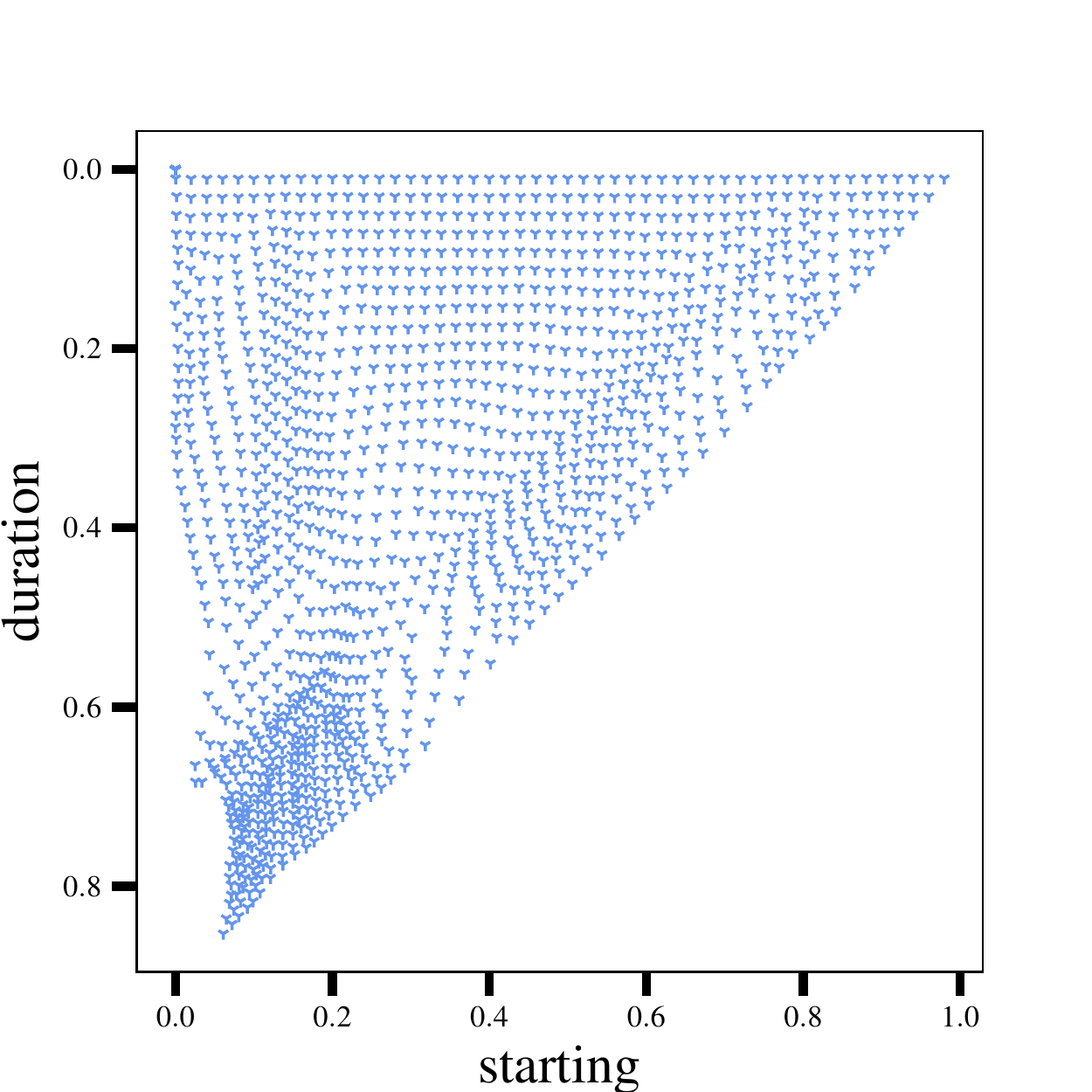}}
\caption{An illustration of the offset flow (a), which is predicted from regular grid, and the updated sample distribution (b).}
\label{fig:offset}
\end{figure}

As shown in Figure~\ref{fig:framework}, the update operator takes base video feature $\mathbf{F}$, confidence maps $\mathbf{G}_m$, and a latent hidden state as input, and outputs updated confidence maps $\mathbf{G}_{m+1}$ and an updated hidden state. 
With each iteration, it produces an update direction $(\bigtriangleup \mathbf{x}_s, \bigtriangleup \mathbf{x}_e)$, and then we perform lookups on the continuous 2D grid (Figure~\ref{fig:offset}. 
These steps are repeated until convergence.
The architecture of our update operator is designed to mimic the steps of the progressive boundary refinement~\cite{pbrnet,tcanet}. 
The update operator is trained to perform refinement such that the sequence converges to a fixed state $\textbf{G}_m \rightarrow \textbf{G}^*$ .

The iterative prediction architecture, following~\cite{raft}, refines the predictions over successive stages, $m \in \{1, . . . , M\}$, with intermediate supervision at each stage. More details are in the supplementary materials. 
Section~\ref{sec:ablation} analyzes the accuracy and generalization for this module.

\subsection{Supervision}
Given a set of ground-truth segment annotations $\mathcal{G}=\left\{\mathbf{g}_n=(\mathbf{g}_{s,n}, \mathbf{g}_{e,n})\right\}_{n=1}^{N}$, the current anchor-based approaches~\cite{bmn,gtad} heavily rely on tIoU scores as the supervisory signal:
\begin{equation}
\left\{\begin{array}{c}
\mathcal{L}_{tIoU} = \mathcal{L}_{bce}(p^1_{\bm{\theta}}, \mathbf{1}\{\text{tIoU}^* > \tau\}) + \lambda_1 \mathcal{L}_{mse}(p^2_{\bm{\theta}}, \text{tIoU}^*)\\
\text{tIoU}^{*} (\mathbf{x}, \mathcal{G}) = \underset{\mathbf{g}_n \in \mathcal{G} } {\max}( \{ \left| \mathbf{x} \cap \mathbf{g}_n\right| / \left| \mathbf{x} \cup \mathbf{g}_n \right| \}),
\end{array}\right.
\end{equation}
where $\tau$ is the front-ground threshold, $p^1_{\bm{\theta}}$ and $p^2_{\bm{\theta}}$ are two type of confidence maps, $\mathcal{L}_{bce}$ is a balanced cross entropy loss,  $\mathcal{L}_{mse}$ is the mean square loss. 
We argue that the tIoU score is actually a non-signed distance~\cite{deepsdf}. 
When each training sample is optimized independently, the network cannot perceive the accurate target location, which leads to a slow convergence~\cite{fineaction}.

Therefore, we add a signed regressing loss as auxiliary supervision signals to predict the time offset for each segment, which shows a better overall performance (see Table~\ref{tab:ablationcr}) .
We adopt the original confidence losses~\cite{bmn} with a boundary regression loss $\mathcal{L}_{offset}$ as below:
\begin{equation}
\left\{\begin{array}{c}
\mathcal{L}_{reg} = \mathcal{L}_{tIoU} + \lambda_2 \mathcal{L}_{offset} \\
\mathcal{L}_{offset} = \left | \triangle \mathbf{x} - (\mathbf{g}^{*}-\mathbf{x})  \right |,
\end{array}\right.
\end{equation}
where $\mathbf{g}^{*}$ denotes the closest ground-truth annotation to the input segment $\mathbf{x}$.

For fair comparisons with our baselines~\cite{bmn,gtad}, we retain the boundary regularization (TEM Loss in BMN~\cite{bmn} and Node Classification Loss in GTAD~\cite{gtad}) and the $\ell$-2 parameter regularization loss:
\begin{equation}
\mathcal{L}_{norm} = \mathcal{L}_{boundary} + \lambda_3 \mathcal{L}_{\ell\text{-}2}(\bm{\theta}).
\end{equation} 

For the iterative process (Section~\ref{sec:rrm}), we use the same loss function, but the truncated return training method is used, and $\alpha$ is given as the attenuation parameter. 
The intermediate supervision at each stage addresses the vanishing gradient problem by replenishing the gradient periodically~\cite{raft}. 
The overall objective is
\begin{equation}
\mathcal{L}_{all} = \sum_{m=1}^{M} \alpha^{m} \mathcal{L}_{reg,m} +\mathcal{L}_{norm}.
\end{equation}

%% file: experiments.tex
\section{Experiments}
\label{sec:experiments}
In this section, we firstly introduce two standard datasets, THUMOS14~\cite{thumos14} and ActivityNet v1.3~\cite{anet}, to evaluate the localization ability and the configuration details of our algorithm.  
Meanwhile, we compare the proposed method, RCL, with existing representative approaches on the two benchmarks.  
Then we carry out the ablation experiments to explore the contribution of each component in our method.  
Finally, we further explore the results on various attributes.

\subsection{Datasets and Evaluation Metrics}
\noindent\textbf{Datasets and features.} We validate our proposed method on two standard datasets: THUMOS14~\cite{thumos14} includes 413 untrimmed videos with 20 action classes. 
According to the public split, 200 of them are used for training, and 213 are used for testing. 
There are more than 15 action annotations in each video; 
ActivityNet v1.3~\cite{anet} is a large-scale temporal action localization dataset with 200 classes annotated. 
The entire 19,994 untrimmed videos are divided into training, validation, and testing sets by ratio 2:1:1.
Each video has around 1.5 action instances.  
To make a fair comparison with the previous works, we use the same two-stream features of these datasets. 
The two-stream features, which are provided by~\cite{tsn}, are extracted by I3D network~\cite{i3d} pre-trained on Kinetics~\cite{kinetics}.  We further validate the effectiveness of our approach with a strong pre-trained feature TSP~\cite{tsp}.

\noindent\textbf{Implementation details.} 
We reimplemented BMN~\cite{bmn} and G-TAD~\cite{gtad} following their respective papers as two discretized baselines. 
We follow the original papers’ training schedules and train our model end-to-end using Adam~\cite{adam} with batch size of 16. 
The learning rate is $6\times 10^{-6}$ on THUMOS14 and $1\times10^{-3}$ on ActivityNet v1.3 for the first 5 epochs, and is reduced by 10 for the following 5 epochs. 
During training, we set weighting parameter $\lambda_1=10, \lambda_2=1, \lambda_3=10^{-5}, \alpha=0.8$ , the front-ground threshold $\tau=0.7$ and training iteration $M=10$. 
During inference, following~\cite{gtad}, we take the segments classification scores from the tIoU and classification branch, and multiply them to produce the proposal score and then fuse our prediction scores with video-level classification scores from~\cite{untrimmednet,tsn}.
For post-processing, we apply Soft-NMS~\cite{softnms}, where the threshold is 0.3 and select the top-$Q$ prediction for final evaluation, where $Q$ is 100 for ActivityNet v1.3 and 200 for THUMOS14.

\noindent\textbf{Metric for temporal action localization.} To evaluate the performance for TAL, we use mean Average Precision (mAP) metric. On THUMOS14 dataset, we report the mAP with multiple tIoUs in set $\{0.3, 0.4, 0.5, 0.6, 0.7\}$. As for ActivityNet v1.3 dataset, the tIoU set is $\{0.5, 0.7, 0.95\}$. 
Moreover, we also report the averaged mAP where the tIoU is from 0.5 to 0.95 with a stride of 0.05.

\begin{table}[t]
\centering
\caption{\textbf{Temporal Action detection results on test set of THUMOS14}, measured by mAP (\%) at different tIoU thresholds. Our RCL achieves the highest mAP for tIoU threshold 0.5 (commonly adopted criteria), significantly outperforming all other methods. 
}
\vspace*{-2mm}
\small
\begin{tabular}{l|ccccc|c}
\toprule
Method     &  0.3 &  0.4 &  0.5 &  0.6 &  0.7  & Short \\ 
    \hline
\multicolumn{7}{c}{End-to-end learned/finetuned on THUMOS for TAL} \\
 \hline

TCN~\cite{tcn}       & - & 33.3 & 25.6 & 15.9 & 9.0 & -\\
R-C3D~\cite{rc3d}    & 44.8 & 35.6 & 28.9 & - & -& -\\
PBRNet~\cite{pbrnet} & 58.5 & 54.6 & {51.3} & {41.8}  & {29.5}& - \\
 \hline
 \multicolumn{7}{c}{Pre-extracted features} \\
 \hline
TAL-Net~\cite{talnet}     & 53.2 & {48.5} & {42.8} & {33.8} & 20.8& -\\
P-GCN~\cite{pgcn}    & {63.6} & {57.8} & {49.1} & - & -& -\\

I.C\&I.C~\cite{icic}& 53.9 & 50.7 & 45.4 & 38.0 & 28.5 & 49.1 \\

 MGG \cite{mgg}   & 53.9 & 46.8 & 37.4 & 29.5 & 21.3& - \\
 BSN~\cite{bsn}   & 53.5 & 45.0 & 36.9 & 28.4 & 20.0& - \\
DBG~\cite{dbg} &  {57.8} & 49.4 & 39.8 & 30.2 & {21.7} & -\\
BMN~\cite{bmn}  & {56.0} & 47.4 & 38.8 & 29.7 & 20.5 & -\\
{G-TAD}~\cite{gtad}    & {54.5} & {47.6} & {40.2} & {30.8} & {23.4} & 44.2 \\
 BC-GNN~\cite{bcgnn}  & 57.1 & 49.1 & 40.4 & 31.2 & 23.1 & -\\
 PBRNet$^{\ast}$~\cite{pbrnet} &54.8 &49.2  &42.3  &33.1   &23.0 &43.6  \\ %
 VSGN~\cite{vsgn}  & 66.7 & 60.4 & 52.4 & 41.0 & 30.4 & 56.6\\
\textbf{RCL (ours)}  &  \textbf{70.1} &  \textbf{62.3} & \textbf{52.9} & \textbf{42.7} & \textbf{30.7} & \textbf{57.1}\\
\bottomrule
\end{tabular}
\raggedright
\footnotesize{
$^*$ Results are referred from ~\cite{vsgn}. They replace 3D convolutions with 1D convolutions to adapt to the feature dimension.}\\
\vspace{-0.3cm}
\label{Tab:exp_thumos}
\end{table}

\subsection{Comparisons with State-of-the-Arts }
\label{sec:sota}
We compare the proposed RCL with recent state-of-the-art methods on the THUMOS14 dataset.
As shown in Table~\ref{Tab:exp_thumos}, with the same pre-trained features, RCL significantly surpasses the grid baseline~\cite{bmn} by absolute $14.1\%$ mAP@0.5, reaching $52.9\%$ mAP@0.5 on THUMOS14.
RCL also demonstrates competitive performance with top-performing temporal action method~\cite{vsgn}, which leverages strong data augmentation and an innovative graph network.
Compared with other iterative optimization method~\cite{pbrnet}, our algorithm has substantially denser samples, which brings $10.6\%$ mAP@0.5 improvement.

\begin{table}[tbp]
\centering
\caption{
\textbf{Action localization results on the validation set of ActivityNet v1.3}, measured by mAPs at different tIoU thresholds and the average mAP. Our RCL, without further finetuning, achieves the state-of-the-art average mAP for most pre-extracted features.
}
\vspace*{-2mm}
\small
\begin{tabular}{l|cccc|c}
\toprule
Method &  0.5  &  0.75  & 0.95 & Average & Short\\
\hline
 \multicolumn{6}{c}{End-to-end learned/finetuned on ActivityNet for TAL} \\
 \hline
CDC~\cite{cdc} & 45.30 & 26.00 & 0.20 & 23.80 & - \\
R-C3D~\cite{rc3d}  & 26.80 & - & - & - & -\\
PBRNet~\cite{pbrnet} & {53.96} & 34.97 & 8.98 & 35.01 & - \\
\hline
 \multicolumn{6}{c}{Pre-extracted I3D~\cite{i3d} features} \\
 \hline
TAL-Net~\cite{talnet} &  38.23 & 18.30 & 1.30 & 20.22 & - \\ 
P-GCN~\cite{pgcn} &48.26 &33.16 &3.27 &31.11 & - \\
I.C \& I.C~\cite{icic}& 43.47&33.91&\textbf{9.21}&30.12 & 14.8 \\
PBRNet$^{\ast}$ ~\cite{pbrnet} & 51.32 &33.33 &7.09 & 33.08 & 17.6 \\
VSGN~\cite{vsgn} & \textbf{52.32}  &35.23   &8.29  & \textbf{34.68} & \textbf{18.8}\\
\textbf{RCL (ours)}  & 51.74 &  \textbf{35.27} & 8.03 & 34.39 & 18.5\\
\hline
 \multicolumn{6}{c}{Pre-extracted TSN~\cite{tsn} features} \\
 \hline
BSN~\cite{bsn}& 46.45  & 29.96 & 8.02  & 30.03  & 15.0\\
BMN~\cite{bmn}&   {50.07} & {34.78} & {8.29} & {33.85}  & 15.2 \\
{G-TAD}~\cite{gtad}  &  {50.36} & {34.60} & {9.02} & { 34.09}  & 17.5 \\
BC-GNN~\cite{bcgnn} & 50.56 & 34.75 & \textbf{9.37} & 34.26 & -\\
PBRNet$^{\ast}$ ~\cite{pbrnet} & 51.41 & 34.35 & 8.66 & 33.90  & 18.0 \\
VSGN~\cite{vsgn} & 52.38 & 36.01  & 8.37 & 35.07 & 19.9 \\
TCANet~\cite{tcanet} &  52.27  & \textbf{36.73}  & 6.86 & 35.52 & - \\
\textbf{RCL (ours)} &  \textbf{54.19}  & 36.19 & 9.17 & \textbf{35.98} & \textbf{20.0} \\
\hline
\multicolumn{6}{c}{Pre-extracted TSP~\cite{tsp} features} \\
\hline
{G-TAD}~\cite{gtad} &   51.26  & 37.12  & \textbf{9.29} & 35.81 & 19.3 \\
VSGN~\cite{vsgn} &   53.26  & 36.76	  & 8.12 & 35.94 & 20.9 \\
\textbf{RCL (ours)} &   \textbf{55.15}  &\textbf{ 39.02}  & 8.27 & \textbf{37.65} & \textbf{21.1} \\
\bottomrule
\end{tabular}
\raggedright
\footnotesize{
$^*$ Results are referred from ~\cite{vsgn}. They replace 3D convolutions with 1D convolutions to adapt to the feature dimension.}\\
 
\label{tab:sota_anet}
\vspace{-0.2cm}
\end{table}

Table~\ref{tab:sota_anet} shows the TAL performace with different features~\cite{i3d,tsn,tsp} on ActivityNet v1.3. 
Among the compared detectors using TSN features, RCL provides the best results with an mAP of $35.98\%$.
RCL achieves a substantial improvement over BMN, with a gain of $1.56\%$ average mAP via TSN features.
Among the compared detectors with TSP~\cite{tsp} features, RCL achieves new state-of-the-art performances with mAP scores of $37.65\%$.
Comparing our RCL with the discretized counterpart~\cite{gtad}, it shows a remarkable gains with $1.8\%$ average mAP, indicating the effectiveness of our detection network under challenging fine-grained scenarios.

\begin{table}[tbp]
\centering
\caption{\textbf{Effectiveness of RCL components on the validation set of ActivityNet v1.3.} CAR is highly effective for short actions. RRM improve the overall performance. }
\vspace*{-2mm}
\setlength{\tabcolsep}{4pt}
\small
\begin{tabular}{ccc|cccc|c}
Baseline & CAR & RRM   &  0.5 & 0.75  & 0.95  & Avg. &Short \\  \shline
\cmark & &   &50.07 & 34.78  &8.02  & 33.85 & 17.5\\ %
\cmark & \cmark &   &52.22  &36.45  & 7.53  & 35.41 & 18.6 \\ %
\cmark & \cmark & \cmark &  \textbf{54.19}  & \textbf{36.19}  & \textbf{9.17} &\textbf{35.98} & \textbf{20.0}\\  %

\end{tabular}
\label{tab:abl_anet}
\vspace{-0.3cm}
\end{table}

\begin{table}[t]\centering
\caption{\bd{Ablation study for the continuous representation on the validation set of ActivityNet v1.3.} 
A naive rescaling leads to a slight decrease in accuracy. 
\label{tab:ablationcr}}
\vspace*{-2mm}
\tablestyle{5pt}{1.0}
\begin{tabular}{@{}c|c|cc|c@{}}
    input dim & output dim & AP@0.5 & mAP & FLOPs (G) \\\shline
100  & 100$\times$100  & 50.07 & 33.85 & 45.6 \\
200 & 200$\times$200  & 51.56 & 33.54 & 91.2 \\
300 & 300$\times$300  & 51.60 & 33.95 & 136.8\\ \hline
100  & 200$\times$200 & 50.79 & 33.11 & 45.6 \\
100  & 300$\times$300 & 50.60 & 33.01 & 45.6 \\ \hline
100  & CAR  w/o offset   & \textbf{52.35} & 34.81  & 63.2 \\
100  & CAR  w/ offset      & 52.22 & \textbf{35.41} & 98.3 \\
\end{tabular}
\vspace{-0.2cm}
\end{table}

\begin{table}[t]\centering
\caption{\bd{Ablation study for the sample strategies on the validation set of ActivityNet v1.3}. 
Our sampling performs better than regular grid sampling and scale-invariant loss~\cite{tsi} indicating the importance of continuous sampling.
\label{tab:ablationsampling}}
\vspace*{-2mm}
\tablestyle{5pt}{1.0}
\begin{tabular}{@{}l|cc|c@{}}
sample strategy & AP@0.5 & mAP & FLOPs (G) \\\shline
regular grid sample   & 50.07 & 33.85 & 45.6 \\
+uniform sample       & 51.60 & 34.14  & 58.8  \\
+scale-invariant sample & \textbf{52.35}& \textbf{34.81} & 63.2 \\ \shline
scale-invariant loss~\cite{tsi} & 51.18 & 34.15 & 45.6 \\
\end{tabular}
\vspace{-0.2cm}
\end{table}

\begin{table}[t]\centering
\caption{\bd{Ablation study for recurrent refine module on the validation set of ActivityNet v1.3.} 
\label{tab:ablationiter}}
\vspace*{-0.2cm}
\tablestyle{5pt}{1.0}\begin{tabular}{c|c|cc@{}}
Update & backbone & AP@0.5 & mAP \\\shline
 & TSN & 52.35 \phantom{+1.84} & 35.41 \phantom{+0.57}\\
 \cmark & TSN & \bd{54.19 \dt{+1.84}} & \bd{35.98 \dt{+0.57}}\\ \hline
  & TSP & 53.76 \phantom{+1.39} & 36.33 \phantom{+0.32}\\
\cmark  & TSP & \bd{55.15 \dt{+1.39}} & \bd{37.65 \dt{+0.32}}\\ 
\end{tabular}
\vspace{-3mm}
\end{table}

\begin{figure*}[t]
\centering
\def\arraystretch{0.9}
\setlength{\tabcolsep}{2.pt}
\begin{tabular}
{lccc}
& BMN~\cite{bmn} & G-TAD~\cite{gtad} & \textbf{RCL(ours)}\\
\mbox{\rotatebox[x=-1.cm]{90}{\rotatebox[x=0.0cm]{270}{(a)}}}&\includegraphics[width=0.32\linewidth]{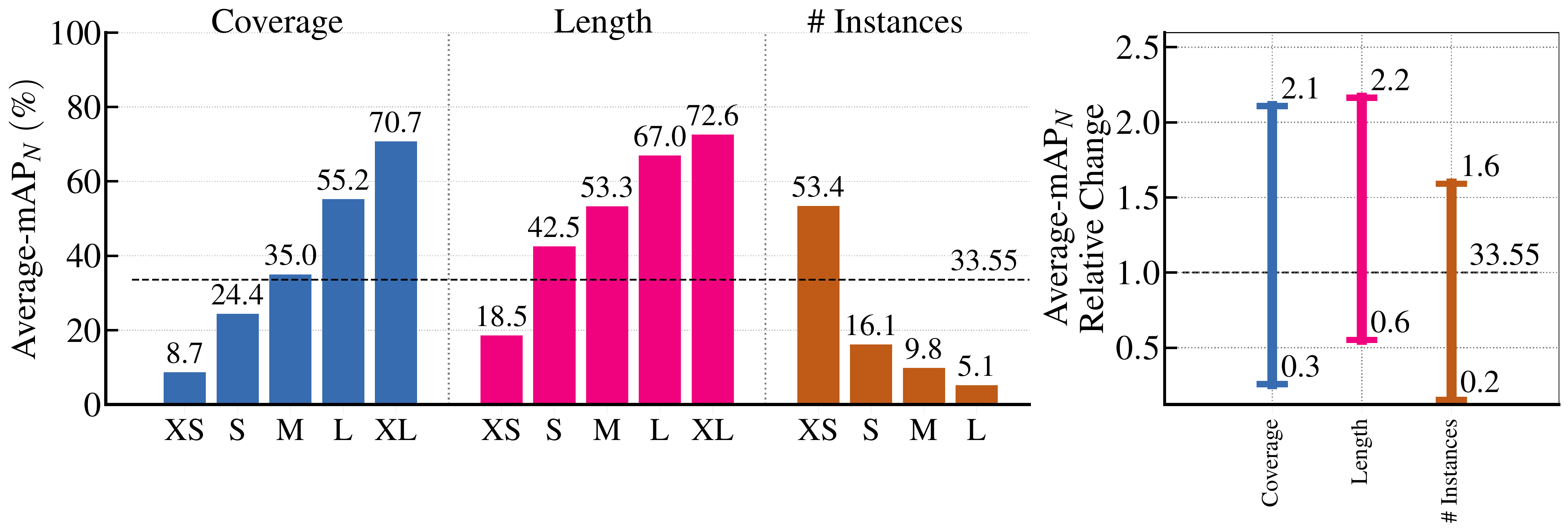}
&\includegraphics[width=0.32\linewidth]{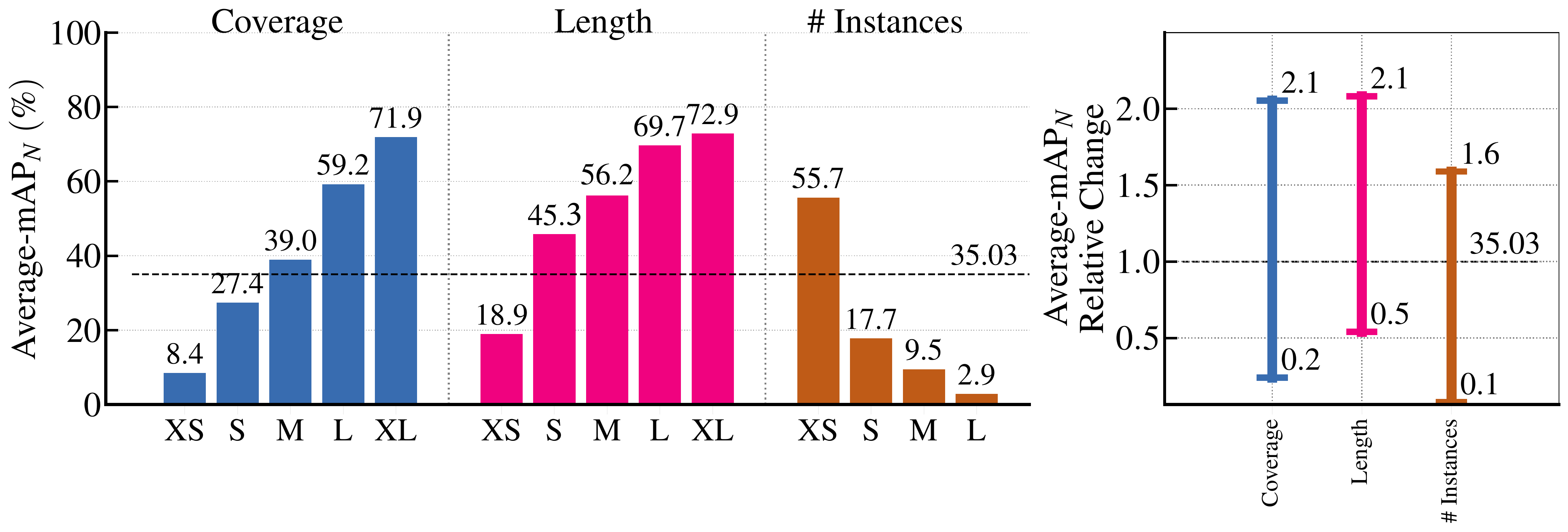}
& \includegraphics[width=0.32\linewidth]{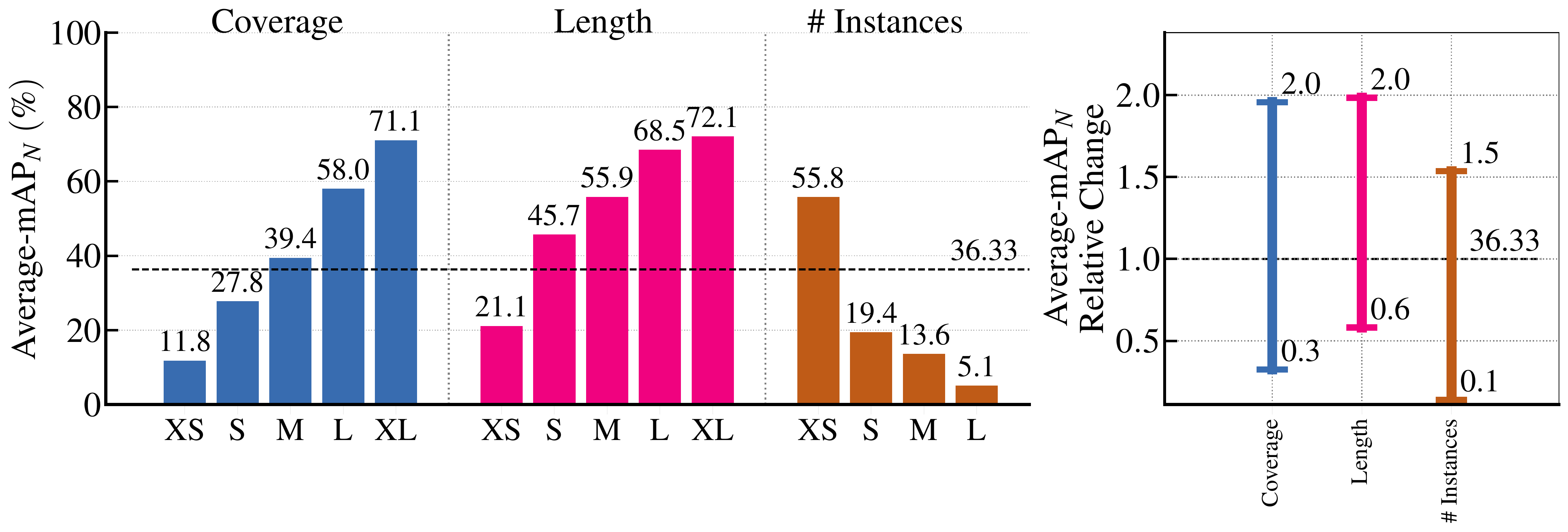}
\\
\mbox{\rotatebox[x=-1.cm]{90}{\rotatebox[x=0.0cm]{270}{(b)}}}& \includegraphics[width=0.32\linewidth]{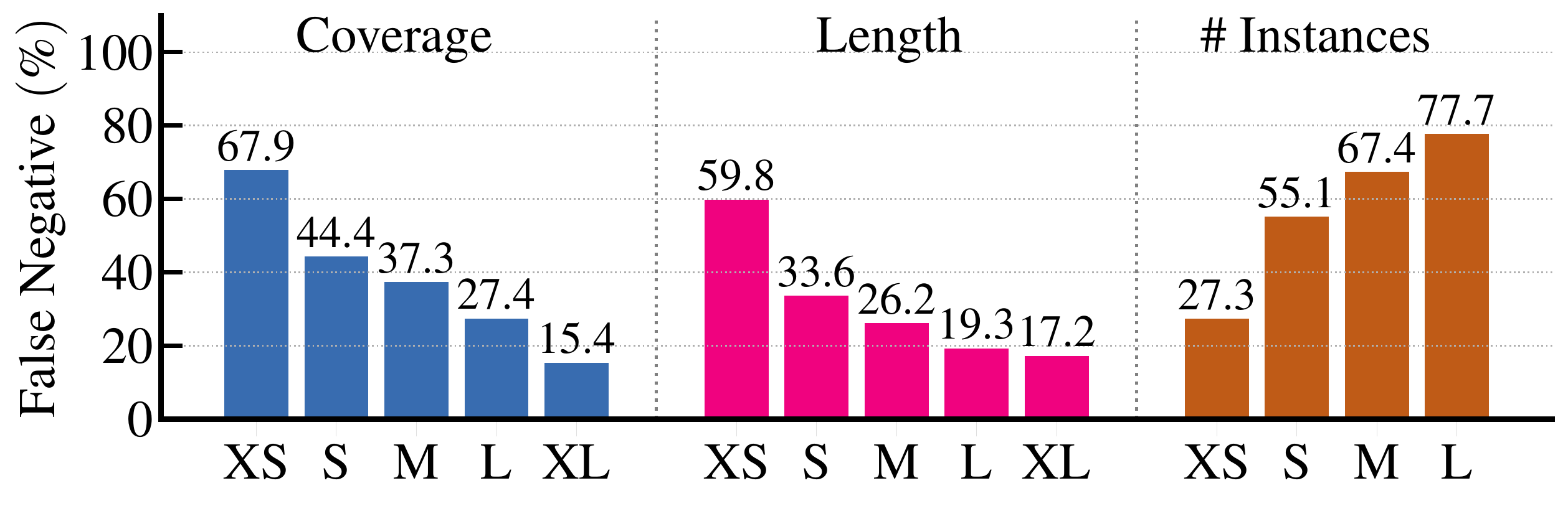}
&\includegraphics[width=0.32\linewidth]{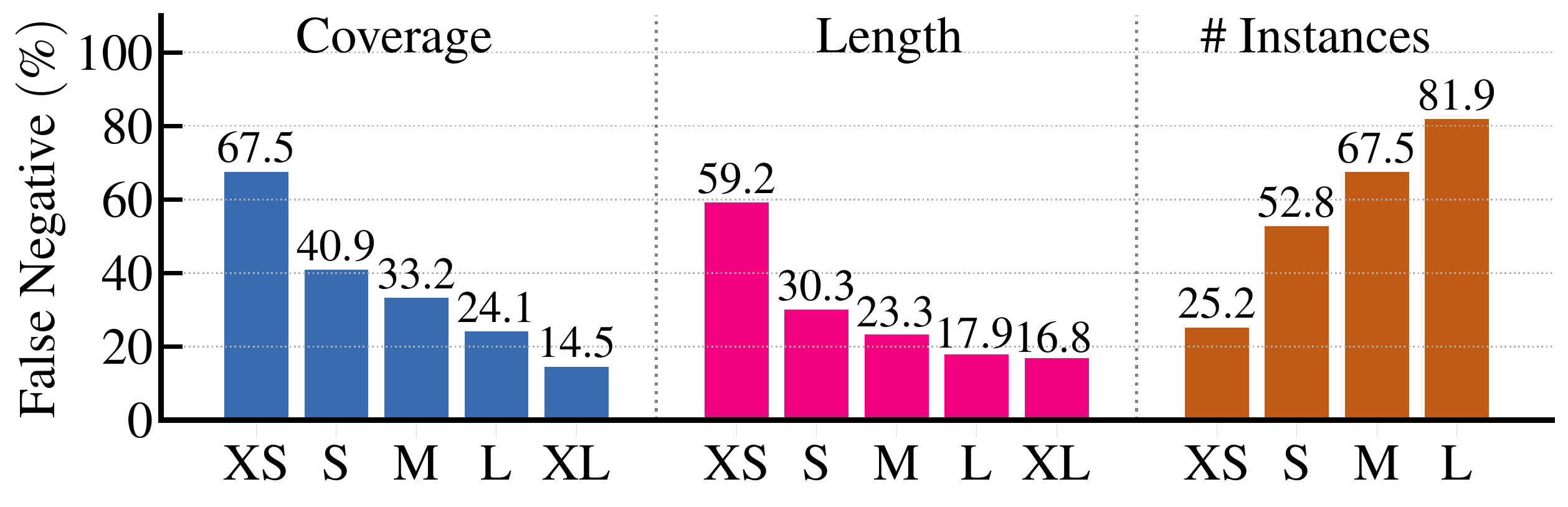}
& \includegraphics[width=0.32\linewidth]{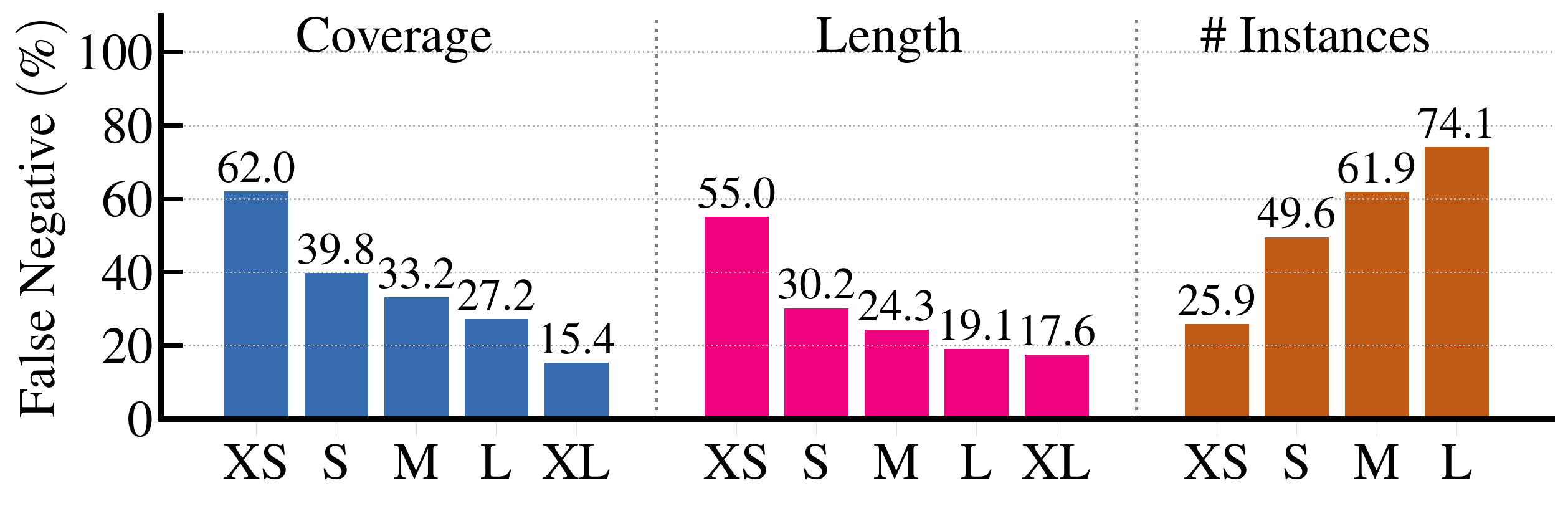}
\\
\mbox{\rotatebox[x=-2.cm]{90}{\rotatebox[x=0.0cm]{270}{(c)}}}&\includegraphics[width=0.32\linewidth]{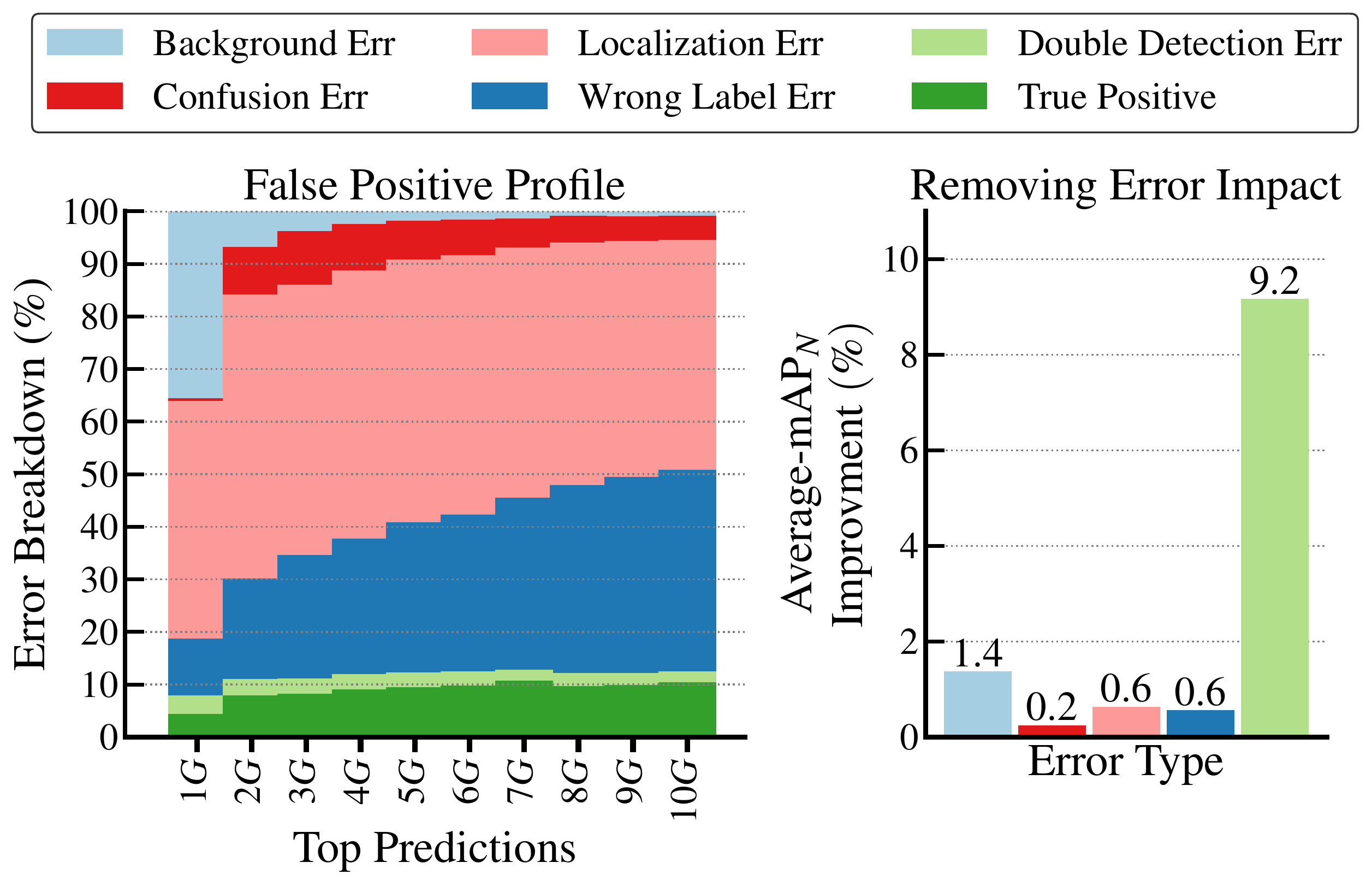}
&\includegraphics[width=0.32\linewidth]{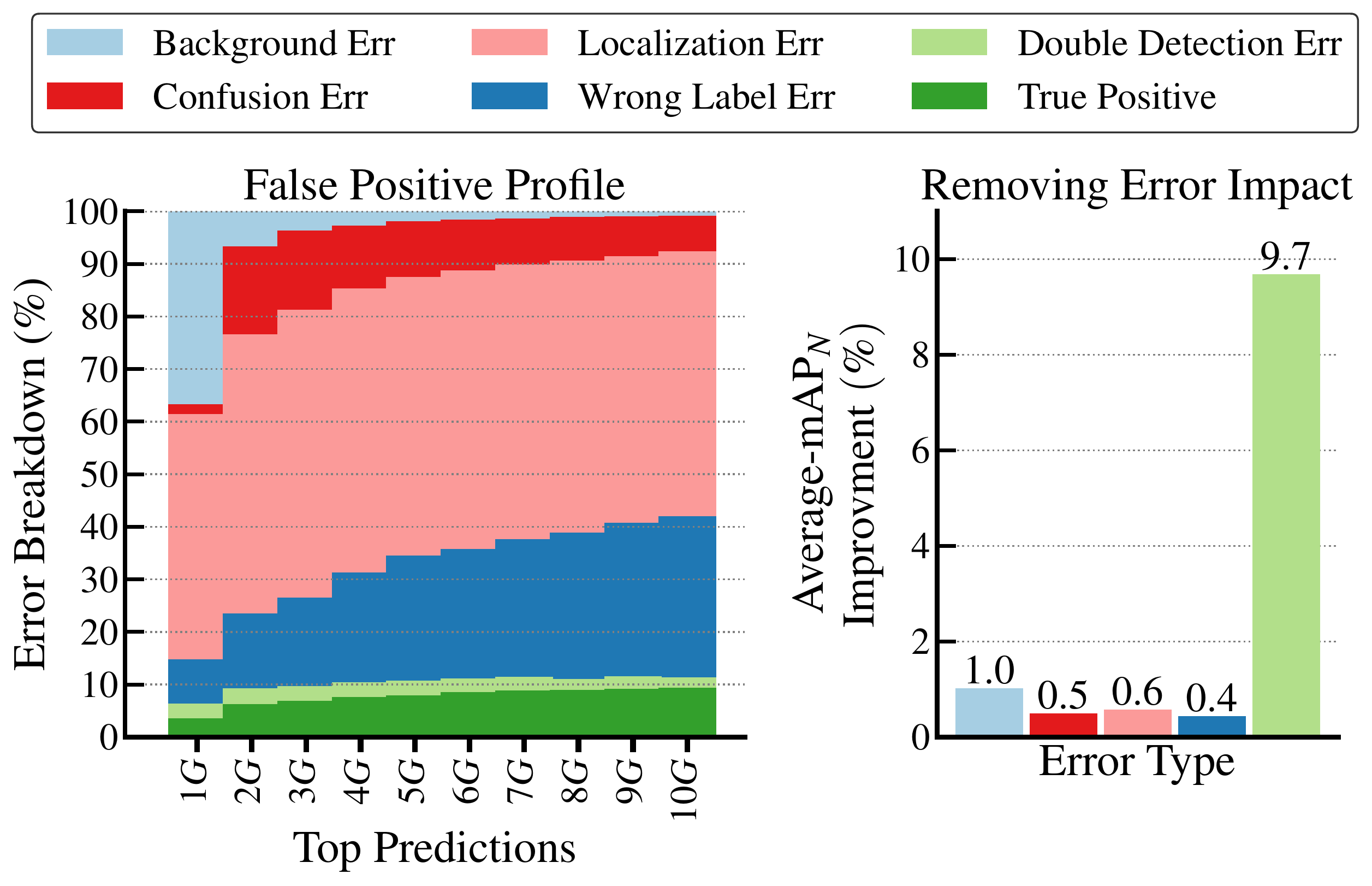}
& \includegraphics[width=0.32\linewidth]{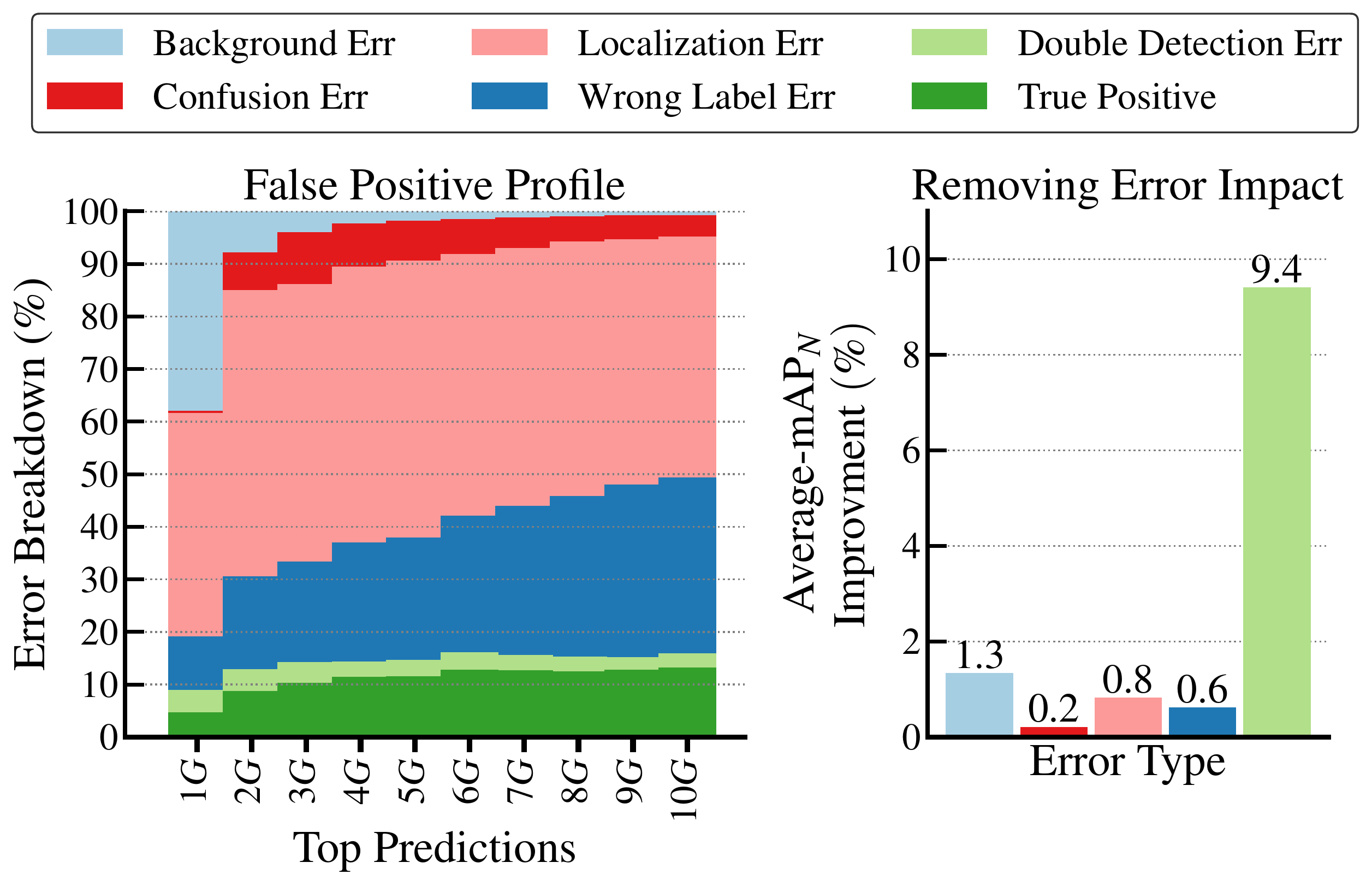}
\end{tabular}
\vspace*{-0.1cm}
\caption{
Illustration of the three types of DETAD analyses~\cite{detad} in ActivityNet v1.3\cite{anet}. (a) The sensitivity  average-mAP$_N$ to action characteristics shows RCL mainly benefits from identifying the tiny segments. (b) The false positive profiles shows RCL significantly reduces the missing detection by $\mathtt{\sim}5.5\%$ for ``Extremely Small'' instances. (c) Average false positive profile  across algorithms for each characteristic.
Actions are divided into five duration groups (seconds): XS (0, 30], S (30, 60], M (60, 120], L (120, 180], and XL (180, inf). Please refer to ~\cite{detad} for more details.}
\label{fig:detad}
\end{figure*}

\subsection{Ablation Study}
\label{sec:ablation}
We evaluate the key components of Continuous Anchoring Representation (CAR) and the Recurrent Refine Module (RRM) with TSN features.
From Table \ref{tab:abl_anet}, we can see CAR obviously improves the performance of short actions as well as the overall performance with $+1.56\%$ average mAP. 
We apply the recurrent module to the refine the heatmaps, which performs $0.57\%$ better than not using the recurrent module. 
This shows that the recurrent optimization indeed helps the model to find the right segments because the initial prediction is a very rough estimation with less context.

To further reveal the devil in the details, a set of simple designs are collected:

\noindent\textbf{The continuous representation:} We first design two naive structures to improve the scale: (1) directly scaling the input size and (2) using bilinear layer to up-sample the final heatmaps. 
As shown in Table~\ref{tab:ablationcr}, we find that mAP is reduced from $33.85\%$ to $33.54\%$ and $33.01\%$, respectively. 
We compare these two upsampling structures to our learned representation and find that the continuous representation cascaded regression significantly help promote the mAP.

\noindent\textbf{Sample strategies:} In Table~\ref{tab:ablationsampling}, we argue that the proposed scale-invariant sample strategy can mitigate the imbalanced data distribution. 
Moreover, instead of directly increasing the weight for a small number of short-term samples, our dense sample strategy can eliminate overfitting and provide a stable estimation for ranking temporal proposals. 

\noindent\textbf{Recurrent refine module:} Table~\ref{tab:ablationiter} shows that the incremental refinements consistently outperform the accuracy on all features.
As a supplement to the offset regression branch, we solve the boundary refinement in an iterative way.

\subsection{DETAD~\cite{detad} Error Analysis}
\label{sec:detad}

To demonstrate the potential gaps with the discretized counterparts, BMN~\cite{bmn} and G-TAD~\cite{gtad} and analyze the sensitivity, 
we show comparisons over three types of DETAD analyses~\cite{detad} on ActivityNet v1.3~\cite{anet} with TSP features~\cite{tsp}.
Figure~\ref{fig:detad} provides meaningful insights for how continuous representation improve the overall performance.  
In Figure~\ref{fig:detad}(a), we can see that the mAP$_N$ is reduced from $72.1\%$ to $21.1\%$ with different segment lengths.
The sharp decline shows that the low detection accuracy of tiny (XS/S) instances is an important bottleneck restricting the overall performance. 
Our RCL consistently outperforms the two baseline methods on tiny instances: Coverage-XS (+3.1/+3.4\%), Coverage-S (+3.4/+0.4\%), Length-XS (+2.6/+2.2\%), Length-S (+3.2/+0.4\%).
This result shows that employing continuous representation is helpful for learning fine-grained clips and thus improves performance.

In addition, Figure~\ref{fig:detad}(b) reveals that RCL achieves the lowest false negative rate with  
Coverage-XS (-5.9/-5.5\%), Coverage-S (-4.6/-1.1\%), Length-XS (-4.8/-4.2\%), Length-S (-3.4/-0.1\%).
The superior performance on false negative profile clearly demonstrates that RCL mitigates the resolution issue of tiny instance and allows to represent shorter segments than other detectors.

Finally, we conduct false positive profile to verify the limitations for our detector and show results in Figure~\ref{fig:detad}(c). 
The most impact error comes from double detection error, which may suffer from the inherent problem in Soft-NMS~\cite{softnms} with low tIoU threshold. 
We hope that a totally end-to-end continuous representation will be a future work.

%% file: conclusion.tex
\section{Conclusion}
\label{sec:conclusion}

In this paper, we propose a continuous representation, which brings a unified perspective for current anchoring representation. 
The proposed representation builds upon an explicit model conditioned with video embeddings and temporal coordinates, which can generate non-uniform anchors of arbitrary length.
We develop an effective scale-invariant sampling strategy and recurrently refine the prediction in subsequent iterations.
The experimental results on the THUMOS14 and ActivityNet v1.3 datasets show the notable performance gain over current state-of-the-art methods, demonstrating that our RCL can detect high fidelity segments.
We hope RCL can serve as a simple yet effective baseline for the community.